\def\year{2021}\relax
\pdfoutput=1
\documentclass[letterpaper]{article} 
\usepackage{aaai21}  
\usepackage{times}  
\usepackage{helvet} 
\usepackage{courier}  
\usepackage[hyphens]{url}  
\usepackage{graphicx} 
\usepackage{graphicx}  
\usepackage{natbib}  
\usepackage{caption} 
\usepackage[switch]{lineno} 
\usepackage{amsfonts}

\title{\textsc{GraphSpy}: Fused Program Semantic-Level Embedding via Graph Neural Networks for Dead Store Detection}
\author{
Yixin Guo,
Pengcheng Li, Yingwei Luo,
Xiaolin Wang, Zhenlin Wang
\\
}
\affiliations{
Peking University, Alibaba Research US, Peking University, Peking University, Michigan Technological University \\
yixinguo@pku.edu.cn,
landy0220@gmail.com,
lyw@pku.edu.cn, wxl@pku.edu.cn, zlwang@mtu.edu
}

\begin{document}

\maketitle

\begin{abstract}
Production software oftentimes suffers from the issue of performance inefficiencies caused by inappropriate use of data structures, programming abstractions, and conservative compiler optimizations. It is desirable to avoid unnecessary memory operations. However, existing works often use a whole-program fine-grained monitoring method with incredibly high overhead.
To this end, we propose a learning-aided approach to identify unnecessary memory operations intelligently with low overhead. By applying several prevalent graph neural network models to extract program semantics with respect to program structure, execution order and dynamic states, we present a novel, hybrid program embedding approach so that to derive unnecessary memory operations through the embedding. We train our model with tens of thousands of samples acquired from a set of real-world benchmarks. Results show that our model achieves 90\% of accuracy and incurs only around a half of time overhead of the state-of-art tool.

\end{abstract}

\section{Introduction}
%
Production software oftentimes suffers from various kinds of performance inefficiencies. Some inefficiencies are induced by programmers during design (e.g., poor data structure selection) and implementation (e.g., use of heavy-weight programming abstractions). Others are compiler induced, e.g., not inlining hot functions. 
Identifying and eliminating inefficiencies in programs are important not only for
commercial developers, but also for scientists writing compute-intensive codes for simulation, analysis, or modeling. Performance analysis tools, such as \emph{gprof}~\cite{Graham+:CC82}, \emph{HPCToolkit}~\cite{Adhianto+:10}, and \emph{vTune}~\cite{VTune:URL}, attribute running time of a program to code structures at various granularities (usage analysis). However, such tools are incapable of identifying unnecessary memory operations (wastage analysis). 

\begin{figure}[t]
\centering
\scalebox{0.95}{
\includegraphics[width=1\linewidth]{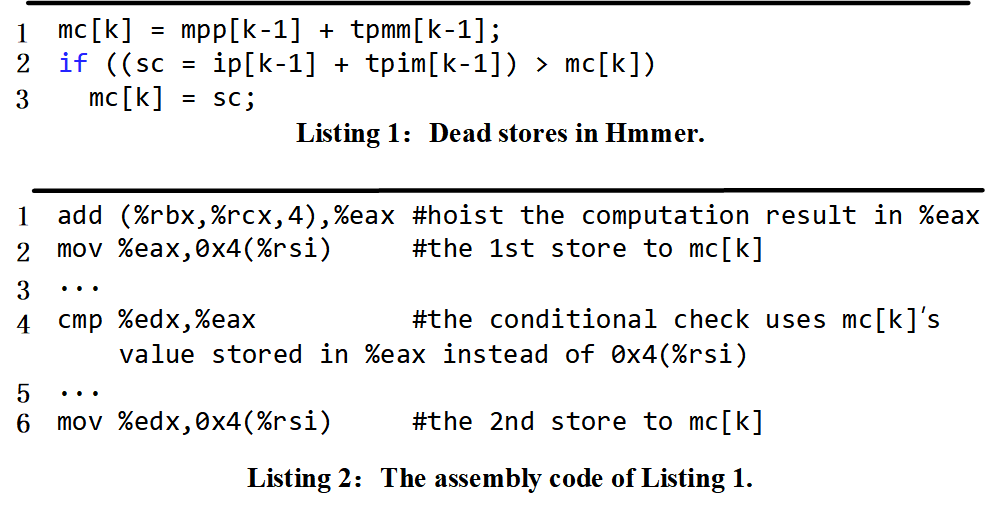}
}
\caption{A dead store example in the \texttt{Hmmer} program.}
\label{fig:dw-ex}
\end{figure}

\emph{Dead stores} are one type of unnecessary memory operations. \emph{A dead store happens when two consecutive store instructions to the same memory location are not intervened by a load instruction for the same location.} Fig.~\ref{fig:dw-ex} shows an example in the \texttt{Hmmer} program from the SPECCPU benchmark suite. The source code first writes, then reads and finally writes \texttt{mc[k]}, showing no dead stores. But in the assembly code in Listing 2, the value of \texttt{mc[k]} is held in a register, which is reused (read) in the comparison on line 4. The memory location \texttt{0x4(\%rsi)} is written first on line 2 and then on line 6. Hence, line 2 is a dead store.

\emph{Whole-program fine-grained monitoring} is a means to monitor execution at microscopic details. It monitors each binary instruction instance, including its operator, operands, and run-time values in registers and memory. A key advantage of microscopic program-wide monitoring is that it can identify redundancies irrespective of the user-level program abstractions. Prior works~\cite{ChabbiM:CGO12,Wen+:ASPLOS17} have shown that the fine-grained profiling techniques can effectively identify dead stores and offer detailed guidance. However, microscopic analysis incurs unacceptable cost in practical use. It is reported that the state-of-the-art (the best-paper awardee) takes up to 150$\times$ run-time slowdown~\cite{Su+:ICSE19}. 


To this end, we propose a learning-aided approach that assists in finding out the procedures that highly likely have dead stores and then employ the fine-grained monitoring tool for these suspicious targets, rather than whole-program fine-grained monitoring. As a result, our approach reduces the running time overhead of the state-of-art tool by 71$\times$.

We formulate this prediction problem as a graph-level task. A procedure is represented by a graph. We embed a procedure through a graph neural network and feed the embedding to a classifier to predict whether dead stores exist. Unfortunately, procedure or program embedding is of an utmost challenge because a program involves many syntactic and semantic factors, including the control flow, data flow, input-sensitivity, context-sensitivity, execution state, architecture-specific instructions, etc. 

The basic building blocks of a procedure are the so-called \emph{basic block}s. By tokenizing a basic block into a set of words, we use word2vec~\cite{Mikolov+:arxiv13} to embed each word and average them as an embedding for representing a basic block. Viewing every basic block as a node, an intra-procedural syntactic structure, i.e. control flow graph (CFG) connects all basic blocks as a graph according to control flow semantics, for example by low-level \texttt{JMP} instructions. In order to capture the intra-procedural structure information, we apply the gated graph neural network~\cite{Li+:ICLR16} (GGNN) onto CFGs through the message-passing neural network framework. 

However, intra-procedural structure information cannot envision dead stores caused across procedure boundaries. In order to make the prediction more precisely, inter-procedural structure information needs to be additionally embedded by dynamically profiling a calling context tree~\cite{Ammons+:PLDI97}. A calling context tree reveals the relations between procedures. We apply the GGNN onto it to capture inter-procedural structure information.

In addition to program structure semantics, we embed dynamic value semantics by taking snapshots of memory addresses and associated values stored during program execution. The snapshots of memory states characterize data dependency between basic blocks within a procedure or procedures, i.e., data flow graph. The embedding of the data flow graph enhances the prediction precision by encoding the input-sensitivity, data flow, and execution state.


Finally, convolutional neural networks are employed to extract common relative positional information between basic blocks in a CFG. It is observed that the relative positional information is common across different architectures and compilation options. With this embedding, our model trained for a specific configuration is capable to detect dead stores
for different platforms and options.

In summary, this paper makes the following contributions:
\begin{itemize}
\item We present a novel approach called \textsc{GraphSpy}, which is inspired by the success of graph neural networks, for detecting dead stores with low overhead. As far as we know, this is the first work that applies GNNs for dead store detection.
\item We present a hybrid embedding approach that embeds both intra- and inter-procedural structure semantics with static CFGs and dynamic calling context trees.  
\item We explore a CNN-based embedding approach to detect dead stores across different micro-architectures and compilation options. 
\item We present an embedding-based data flow graph that captures program value semantics by sampling memory and register states with negligible overhead.
\item We evaluate our approach on a set of real-world benchmarks. The achieved average prediction accuracy is as high as 90\%, and the reduced time overhead doubles the state-of-art monitoring tool.
\end{itemize}

\section{Graph Neural Networks}

\def \GNN {\emph{GNN}}
\def \GGNN {\emph{GGNN}}

The objective of Graph Neural Network is to learn the node representation and graph representation for predicting node attributes or attributes of the entire graph.
A \emph{Graph Neural Network (GNN)}~\cite{Scarselli+:ITNN09} structure $G = (V, E)$ consists of a set of node $V$ and a set of edge $E$. 
Each node $v \in V$ is annotated with an initial node embedding 
by $x \in \mathbb{R}^D$ and a hidden state vector $h_v^t \in \mathbb{R}^D$ ($h_v^0$ often equals to $x$). 
A node updates its hidden state by aggregating its neighbor hidden states and its own state at the previous time step. In total, $T$ steps of state propagation are applied onto a GNN. 
In the $t$-th step, node $v$ gathers its neighbors' states to an aggregation as $m_v^t$, as shown in Eq.~\ref{eq:aggr}. Then the aggregated state is combined with node $v$'s previous state $h_v^{t-1}$ through a neural network called $g$, as shown in Eq.~\ref{eq:upd}. $f$ can be an arbitrary function, for example a linear layer, representing a model with parameters $\theta$. \begin{equation}
\label{eq:aggr}
m_v^t = \sum_{(u, v) \in E} f(h_v^t; \theta)
\end{equation}
\begin{equation}
\label{eq:upd}
h_v^{t} = g(m_v^t; h_v^{t-1})
\end{equation} 


\emph{Gated Graph Neural Network (GGNN)}~\cite{Li+:ICLR16} is an extension of GNN by replacing $g$ in Eq.~\ref{eq:upd} with the \emph{Gated Recurrent Unit (GRU)}~\cite{Chung+:arxiv14} function as shown in Eq.~\ref{eq:upd-gru}. The GRU function lets a node memorize history long-term dependency information, as  it  is  good  at  dealing  with long sequences by propagating the internal hidden state additively instead of multiplicatively.
\begin{equation}
\label{eq:upd-gru}
h_v^{t} = \textbf{GRU}(m_v^t; h_v^{t-1})
\end{equation}

Following GGNN, quite a few different extensions to GNN were developed by applying various deep learning techniques. \emph{Graph convolution network (GCN)}~\cite{KipfW:CoRR16} was proposed by adding convolutional layers to update node embeddings. \emph{Graph attention network (GAT)}~\cite{Velickovic+:arxiv18} leverages the attention mechanism to formalize the spatial and time sequential information. \emph{GraphSAGE}~\cite{Hamilton+:NIPS17} adopts an aggregating function to merge the node and its neighbor nodes. For better graph representation learning, two prevalent frameworks \emph{Message Passing Neural Network} (MPNN)~\cite{Gilmer+:corr17} and \emph{Graph Network} (GN)~\cite{Battaglia+:corr18} were proposed. MPNN has a message passing phase and a readout phase. The message passing phase runs several steps to capture the information from neighbor nodes. The readout phase computes an embedding for the whole graph. 


\section{Program Representation and Embedding}
\label{sec:appr}
\begin{figure*}[t]
\centering
\scalebox{0.8}{
\includegraphics[width=1.1\linewidth]{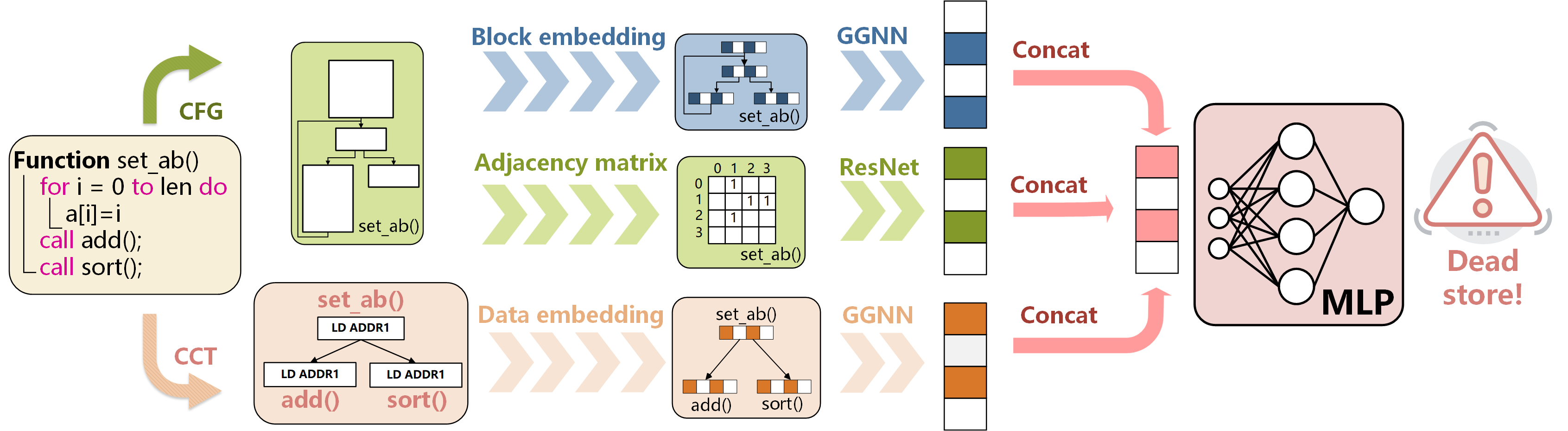}
}
\caption{Overview of the program representation using fused static and dynamic embeddings based upon GNN.}
\label{fig:arch}
\end{figure*}

We now present the overview of the proposed approach. First, provided a procedure we construct a CFG from its binary code and employ a static structure-aware embedding approach to encode the intra-procedural structure semantics. Second, we  profile run-time calling context trees to embed inter-procedural structure semantics and dynamically take program snapshots to capture memory states including register values and memory values to embed value semantics. Finally, a CNN model is employed to embed relative positional information for a CFG. 
Putting together these embeddings as a whole embedding, we predict if the target procedure exists dead stores. Fig.~\ref{fig:arch} shows the diagram of the overview of the proposed approach using fused static and dynamic embedding.

\subsection{Static Intra-Procedural Structure Embedding}

\begin{figure*}[t]
\centering
\scalebox{0.8}{
\includegraphics[width=1\linewidth]{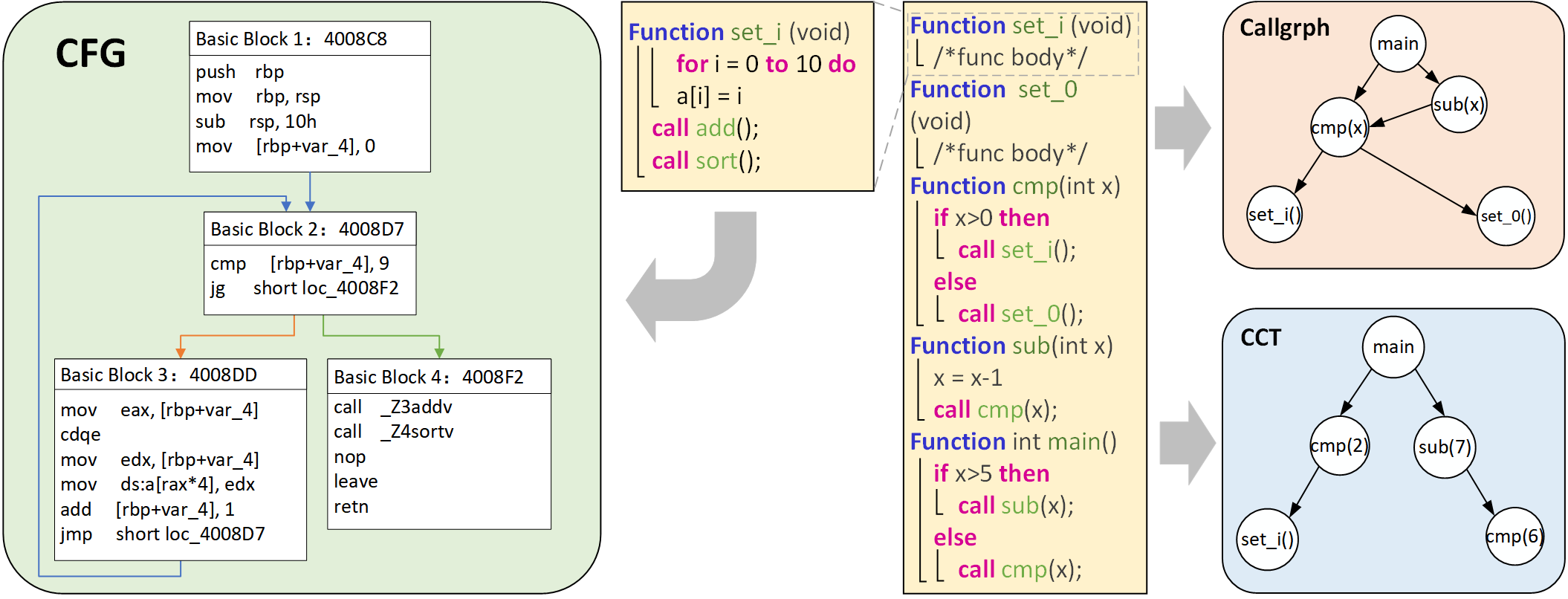}
}
\caption{Examples of a control-flow graph and the comparison between a call graph and a calling-context tree.}
\label{fig:cc-tree}
\end{figure*}

\def \CFG {\emph{CFG}}

An \emph{assembly code} can be compiled from a source code. An assembly code is specific to a processor architecture, for example x86-64. It is a flat profile of \emph{instruction}s, $Ins_i (i \in 1 \dots m)$. Each instruction $Ins_i$ is composed of a series of \emph{token}s, $t_i (i \in 1 \dots n)$. These tokens are of many types, including operators and operands. Some instructions are compute operations with respect to register values (e.g., \texttt{ADD}, \texttt{SUB}). Some instructions move values between registers and memory (e.g., \texttt{LOAD}, \texttt{STORE}). Others represent conditional branch or jump to other locations (e.g., \texttt{JZ, JMP}).

A basic block $B_i (i \in 1 \dots K$) is composed of a sequence of instructions without any control flow statements. It has only one entry and one exit statement. When the program pointer jumps from a source to a target block, we connect an edge from the source to the target. Viewing the basic blocks as nodes and the connections as edges, we form a \emph{control flow graph (CFG)}. For instance, for x86 direct branches, there are only two possible target blocks for a given source block, which we can refer to as the \emph{true block} and \emph{false block}. Fig.~\ref{fig:cc-tree} shows a procedure named \texttt{set\_i} and its according CFG, where it is observed that each basic block has only one entry and one exit and a basic block may have multiple following basic blocks.

A \CFG\ represents intra-procedure program structure semantic, so we attempt to embed a \CFG\ into a feature map. First, we apply the word2vec~\cite{Mikolov+:arxiv13} model to encode every token in an instruction inside a basic block. Then we average out token embeddings as an embedding of an instruction, and finally derive a whole embedding of a basic block by averaging the embeddings of instructions. Eq.~\ref{eq:cfg-em} depicts the embedding process.
\begin{equation}
\label{eq:cfg-em}
\forall k, e_{bb_k} = \frac{\sum_{j=0}^m \{e_{Ins_j} = \frac{\sum_{i=0}^n \textbf{word2vec}(t_i)}{n}\}}{m}
\end{equation}

Once we have embedded every basic block, we formalize a graph of embeddings. We start to run the GGNN model through the message-passing neural network model. The aggregate function for each node to compute the state vector is computed using a gated recurrent unit (GRU), as shown in Eq.~\ref{eq:upd-gru}. GRU is especially good at capturing long-time dependency between the executions of basic blocks within a procedure. We embed it from the assembly without any actual executions, therefore this type of embedding is static and takes no cost for the run-time execution.



\emph{Abstract Syntax Tree} (AST)~\cite{Neamtiu+:SEN2005} is an intermediate representation of a binary generated in the process of compilation from the source code to binary code. It uses context-free grammar parsing rules that partially include the grammars and execution orders of a binary. 
By contrast, we choose to construct a \CFG\ from the assembly code, because one of the benefits is that it is typically less stylish and tighter to program semantics. For example, programs that are syntactically different but semantically equivalent tend to correspond to similar assembly codes. Moreover, the assembly code often embrace more architecture-specific characteristics than AST so that the embedding encodes architecture level information.

\subsection{Dynamic Inter-Procedural Structure Embedding}


\def \CCT {\emph{CCT}}

CFGs represent only intra-procedural structure semantic, rather than whole-program structure. \emph{Call graph} captures inter-procedural structure information, i.e., caller-callee relations. Nevertheless, call graph only expresses static structure information that may not reflect actual  calling sequence of function calls. An alternative is to profile \emph{calling context tree (CCT)}s~\cite{Ammons+:PLDI97}, which is the actual function-call sequence during program execution.

In \emph{CCT} profiles, data (e.g., function call) is collected with respect to each stack frame. \CCT s comprise the merged call-stacks from all functions invoked during the course of program execution, with each frame in the stack represented as a node in the tree, and with common prefixes merged. Typically, each call stack in the \CCT\ is rooted at \texttt{main}.  A \CCT\ can differentiate a function called in different ways from different contexts with different parameters and global states. 
Furthermore, when time measurement data is aggregated across threads, the aggregated time obscures problems occurring on a specific subset of \CCT s.

Fig.~\ref{fig:cc-tree} compares the \emph{static} call graph and \emph{dynamic} calling context tree constructed from the same code. In the upper call graph, there is only one instance for any single procedure and the caller-callee relations are formed statically, for example from \texttt{cmp(x)} to \texttt{set\_0()}. While in the bottom \CCT\  example, two instances of \texttt{cmp()} exist with different input actual parameters, one of which is called by \texttt{sub(7)} and the other calling \texttt{set\_i()}.

In addition to the \CCT\ profiles, we periodically take program snapshots during program execution which are sets of values that occur over the course of the execution. We call every program snapshot, i.e. a set of values, as a program memory state, which is defined as a set of general-purpose registers and memory addresses and associated values. Different than registers, memory addresses are indexed by a 64-bit integer and hence extremely widely spread. In theory, we should include all values of registers and entire memory address space for a program memory state. However, it takes unacceptable space and time cost to profile entire memory address space. Instead, for a program snapshot, we record general-purpose register values and only memory values stored within this execution. These values are lightly obtainable with negligible overhead via a binary instrumentation tool called DynamoRIO~\cite{dynamorio:2004}.

We associate program memory states to every function instance in \CCT\ profiles, thus each function instance has a set of memory-state related values. As embedded for every basic block in the previous section, we use the word2vec model to embed the memory values of a function instance in a \CCT\ profile. After that, we run the GGNN model again on the \CCT\ graph to propagate memory state semantics across procedure boundaries to formalize a \CCT\ embedding.

Program memory states characterize memory value semantics, i.e., which functions load and store which memory addresses. Because dead stores may happen between two function instances, the GNN propagation of memory states in the \CCT\ aides in capturing these dead store cases. We call the embedding of a \CCT\ as inter-procedural structure semantics and memory value semantics.

\subsection{Relative Positional Information Embedding}
It is always desirable to train a generalized model for different micro-architectures and compilation options. 
\emph{Convolutional neural network}s (CNNs) are leveraged to learn relative positional information between basic blocks in a CFG. It is observed that the relative positional information between basic blocks are commonly shared by different micro-architectures and compilation options.

\begin{figure}[t]
\centering
\scalebox{0.7}{
\includegraphics[width=1.1\linewidth]{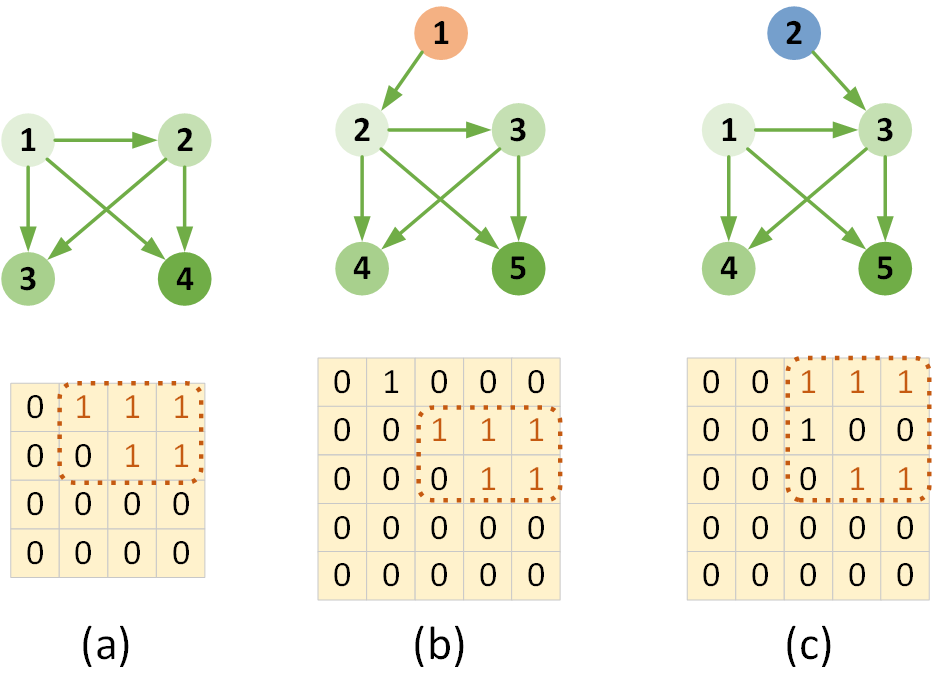}
}
\caption{Relative positional modeling.}
\label{fig:order}
\end{figure}

Fig.~\ref{fig:order} shows three CFGs and their according adjacency matrices, which could be converted mutually by making minor changes. Each of the three CFGs has a rectangle, showing a relative positional relation of four basic blocks. As is shown, the rectangle framed share similarities, i.e., the numeric matrix of this rectangle in (a) is (1, 1, 1, 0, 1, 1). Interestingly, this numeric relation invariance holds for different architectures and options. Consider (b) and (c) as the resulting adjacency matrices for different architectures and options. 
In (b), a new node is added into its adjacency matrix, but the spatially relative positions of those nodes of the corresponding rectangle stay the same and so does its numeric matrix. Thus, the numeric relation invariance holds for the transformation.
In (c), the added node seems breaking the numeric matrix. But when removing the second row, the numeric matrix stays the same as (1, 1, 1, 0, 1, 1). That said, the numeric relation invariance holds. It is just like capturing the spatial information in a image, so we use CNNs.

CNN may learn this type of numeric relation invariance. When building binaries with different options for different architectures, the relative positional relations between basic blocks are oftentimes similar. 
By CNN embedding, our model desires to exploit a one-fits-all solution. Additionally, using CNN can save time overhead since the computation of CNNs upon adjacency matrices is a lot faster comparing to traditional graph feature extraction algorithms. Moreover, CNNs could be added to inputs with varying sizes, so it is able to model graphs of different sizes without pre-processing such as padding.
\begin{equation}
\label{eq:resnet}
g_{\mathcal{A}} = \textbf{Maxpooling} (\textbf{Resnet} (\mathcal{A}))
\end{equation}

As shown in Eq.~\ref{eq:resnet}, we apply the Resnet~\cite{He+:CVPR16} model onto the adjacency matrices to encode the relative positional information. A MaxPooling layer is added after the last layer of the Resnet network to reduce high dimensional data to low dimension. We use an 11-layer Resnet with 3 residual blocks. 
$\mathcal{A}$ denotes an adjacency matrix. $g_{\mathcal{A}}$ is the final embedding.

\subsection{Fused Static and Dynamic Embedding}
Putting together all aforementioned embeddings, a single final program embedding is derived for the target procedure (Eq.~\ref{eq:final}). Finally, we run the embedding through an MLP to check if the target procedure exists dead stores (Eq.~\ref{eq:output}). Fig.~\ref{fig:arch} summarizes the overview of the entire approach.
\begin{equation}
\label{eq:final}
g_{prog} = \textbf{Dense} (g_{\mathcal{A}} \mathbin\Vert g_{intra} \mathbin\Vert g_{inter}) 
\end{equation}
\begin{equation}
\label{eq:output}
g_{o} = \textbf{MLP} (g_{prog})
\end{equation}

When the target procedure is predicted to be suspicious, we apply the state-of-art fine-grained monitoring tool to carefully scan the target procedure line-by-line. As a result, the tool outputs the memory locations where it assures that dead stores happen. In this way, our model gets rid of a number of procedures that do not have dead stores, therefore, a substantial amount of time and space overhead for checking these procedures may be saved.

\section{Evaluation}
In this section, we answer the following questions:
\begin{itemize}
\item What is the prediction accuracy by our model for any single machine and compilation configuration?
\item What is the cross-platform prediction accuracy of a unified model targeting at different architectures and  options?
\item What is time and space overhead improvement in contrast to the state-of-the-art fine-grained monitoring tool?
\item What is the performance comparison of the proposed approach to other popular models?
\end{itemize}

\subsection{Experimental Setup}
\noindent
\textbf{Dataset.} The proposed approach is composed of two types of inputs, static assembly code and dynamic program profiles. In order to collect static assembly, we compile source code for a binary through the \texttt{GCC} compiler and then disassemble the binary by using GNU binary utilities.

\begin{table}[t]
\centering
\scalebox{0.9}{
\begin{tabular}{c|c|c|c}
\hline
\textbf{Binary} & \textbf{\#Functions} & \textbf{\#BBs} & \textbf{Data Volume} 
\\\hline
505.mcf & 67 & 799 & 2.26GB  \\
508.namd & 92 & 1199 & 460MB  \\
510.parest & 18694 & 275693 & 10.51GB    \\ 
520.omnetpp & 7451 & 56217 & 4.08GB \\
523.xalancbmk & 12613 & 145245 & 4.33GB \\
526.blender  & 36782 & 314951 & 31.91GB  \\
544.nab & 287 & 7030 & 259MB  \\
557.xz & 355 & 4685 & 813MB \\ 
\end{tabular}
}
\caption{Benchmark statistics.}
\label{tbl:rawdata}
\end{table}

Dynamic profiles are obtained for calling context trees and memory-state values using the binary instrumentation tool DynamoRIO~\cite{dynamorio:2004}. As is all known SPECCPU benchmark suite~\cite{SPEC-CPU} is a golden standard used to evaluate software system performance. We run integer programs from the benchmark suite with the reference input and sample program execution with for a few times. Each sample records one million instructions continuously. Our tool attaches to the running process, fast forwards to the regions of interest and outputs desired profiling memory-state data. Table~\ref{tbl:rawdata} shows the statistics of programs used, including the number of functions, number of basic blocks and, more importantly, raw data file volume that contains raw data of instructions, CFGs, memory values, CCT profiles, and label data.

\vspace{+.3em}
\noindent
\textbf{Tools.} 
For dataset collection, two profiling tools were implemented and one existing tool is utilized. In order to obtain the ground truth data, i.e., dead stores of all binaries, we made our best efforts to carefully implement the state-of-the-art fine-grained monitoring tool, \texttt{CIDetector}~\cite{Su+:ICSE19} based on DynamoRIO. \texttt{CIDetector} monitors every memory load and store instruction. We attribute all found dead stores to every procedure of every binary. Each procedure corresponds to a data sample since we are predicting on a per-procedure basis. The procedure is labeled as 1 (true) if it has dead stores, 0 (false) otherwise.

A binary analysis tool called \texttt{angr}~\cite{Shoshitaishvili+:ISSP16} is utilized to construct CFGs by taking as input the assembly code and outputting a CFG for every single procedure. A simple profiling tool was implemented to dump calling context trees and memory states based on DynamoRIO.

By distinguishing two micro-architectures, \emph{x86-64} and \emph{ARM} (AArch64) and compiling with two separate options, \texttt{GCC} \emph{-O2} and \texttt{GCC} \emph{-O3}, we compose four configurations: \emph{x86-O2}, \emph{x86-O3}, \emph{ARM-O2}, and \emph{ARM-O3}. We shuffle the samples from all binaries with any single configuration and take a fraction as training and testing datasets, as shown in Table~\ref{tbl:dataset}. The size ratio of true samples over false samples is around 1:1, for a balance of distribution.

Additionally, we create a hybrid configuration by mixing samples from four basic configurations. This configuration is used to train a unified model that aims to work for any compilation option or micro-architecture.

\begin{table}
\centering
\begin{tabular}{c|c|c|c|c}
\hline
\textbf{Config.} & \textbf{Training} & \textbf{Validation} &  \textbf{Test} & \textbf{Total}
\\\hline
{x86-O2} & 13,999 & 13,781 & 13,782 & 41,562 \\
{x86-O3} & 13,271 & 13,660 & 13,660 & 40,591 \\
{ARM-O2} & 13,633 & 13,633 & 13,063 & 39,760 \\
{ARM-O3} & 13,408 & 13,136 & 13,138 & 39,682 \\
Hybrid   & 40,978 & 28,320 & 28,323 & 97,621
\end{tabular}
\caption{Dataset sizes of different configurations.}
\label{tbl:dataset}
\end{table}


\vspace{+.3em}
\noindent
\textbf{Alternative models.} 
BERT~\cite{Vaswani+:corr17} makes use of the attention mechanism to learn contextual relations between words in a sentence. BERT is an alternative to word2vec to encode a basic block and memory states. CNN-based models are alternatives to Resnet to encode relative positional information. We
measure the accuracy results of using word2vec, Resnet (7 or 11 layers), BERT, or CNN-based models alone and the accuracy results by combining some of them, and make a comparison with \textsc{GraphSpy}.

\vspace{+.3em}
\noindent
\textbf{Hyper-parameters.} 
We show only the values of all hyper-parameters used in the final experiments. The learning rate used is 0.0001 and the batch size is 64. We use the word2vec models in two places for basic block embedding and memory value embedding. The dimension of the former embedding is 60 and the latter 30. GGNN is also used for CFGs and CCTs. The output dimension and number of steps are 70 and 10 for CFGs, and 50 and 5 for CCTs. The output dimension of Resnet is 40. We also tuned these parameters for a wide range, which will be given in Appendix.

\vspace{+.3em}
\noindent
\textbf{Platform.} 
All deep learning tasks were performed on 8 Nvidia Tesla V100 GPU~\cite{nvidia:v100} cards of the Volta architecture. Each has 5120 streaming cores, 640 tensor cores and 32GB memory capacity. The CPU host is Intel Xeon CPU 8163 2.50GHz, running Linux kernel 5.0. All the other non-deep learning tasks were run on the host.

\subsection{Prediction Results}
\begin{figure}[t]
\centering
\scalebox{0.8}{
\includegraphics[width=1\linewidth]{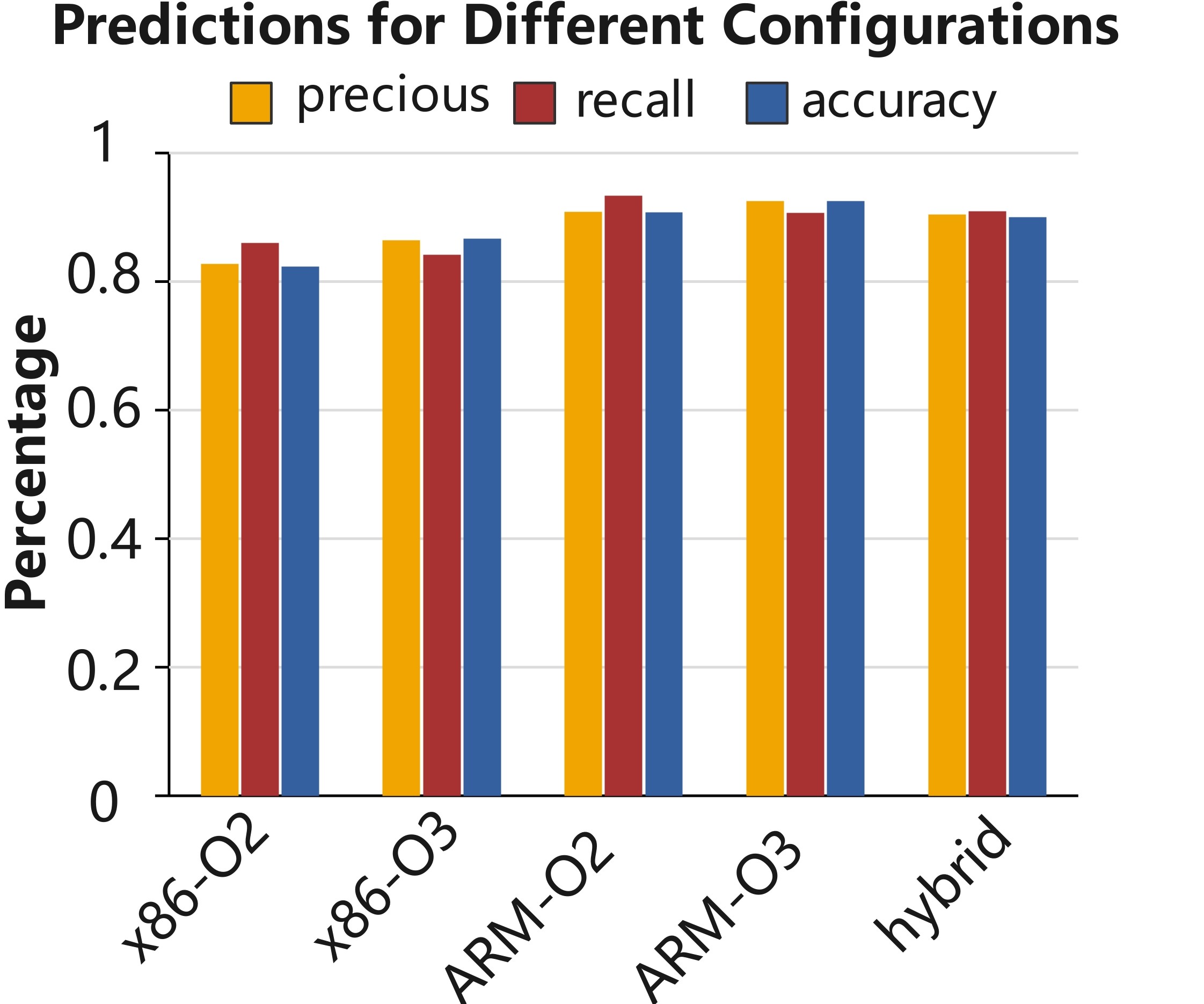}
}
\caption{Prediction results for different and hybrid configurations.}
\label{fig:pred-result}
\end{figure}

Fig.~\ref{fig:pred-result} measures three metrics, precision, recall rate and accuracy for each configuration and the hybrid configuration. The precision metric is computed as $\frac{TP}{TP + FP}$, the recall rate as $\frac{TP}{TP+FN}$, and the accuracy as $\frac{TP + TN}{TP+FN+FP+TN}$, where $TP$ denotes true positive, $FP$ false positive, $FN$ false negative, and $TN$ true negative. Both the precision and recall rate metrics demonstrate the capability of picking just the true samples from all samples. 

Clearly, all measurements are beyond 80\%, which just confirms the efficacy of the proposed model for different configurations. On average, we achieve 88.14\% of precision, 88.55\% of recall rate and 88.08\% of accuracy, respectively. In particular, the results for \emph{ARM} is uniformly better than those for \emph{x86}. The high prediction accuracy, together with 50-50 distributed true samples and false samples, implies that the proposed model is capable to rule out false samples for dead store detection. Hence, we may save up to half of the checking overhead by~\texttt{CIDetector} for omitting those samples (i.e. procedures). We will present memory overhead reduction and time saving in the next section.

It is worth noting that we achieve as high as accuracy of 90.02\% in the hybrid configuration. We trained the model with samples from different architectures and options, and test for samples from different configurations as well. Results demonstrate that the proposed model has a great potential to be a unified model that performs well across different architectures and compilation options, therefore we would argue that the proposed model can be a one-fits-all solution.

\subsection{Overhead Results}
\begin{figure*}[t]
\centering
\scalebox{0.8}{
\includegraphics[width=1.1\linewidth]{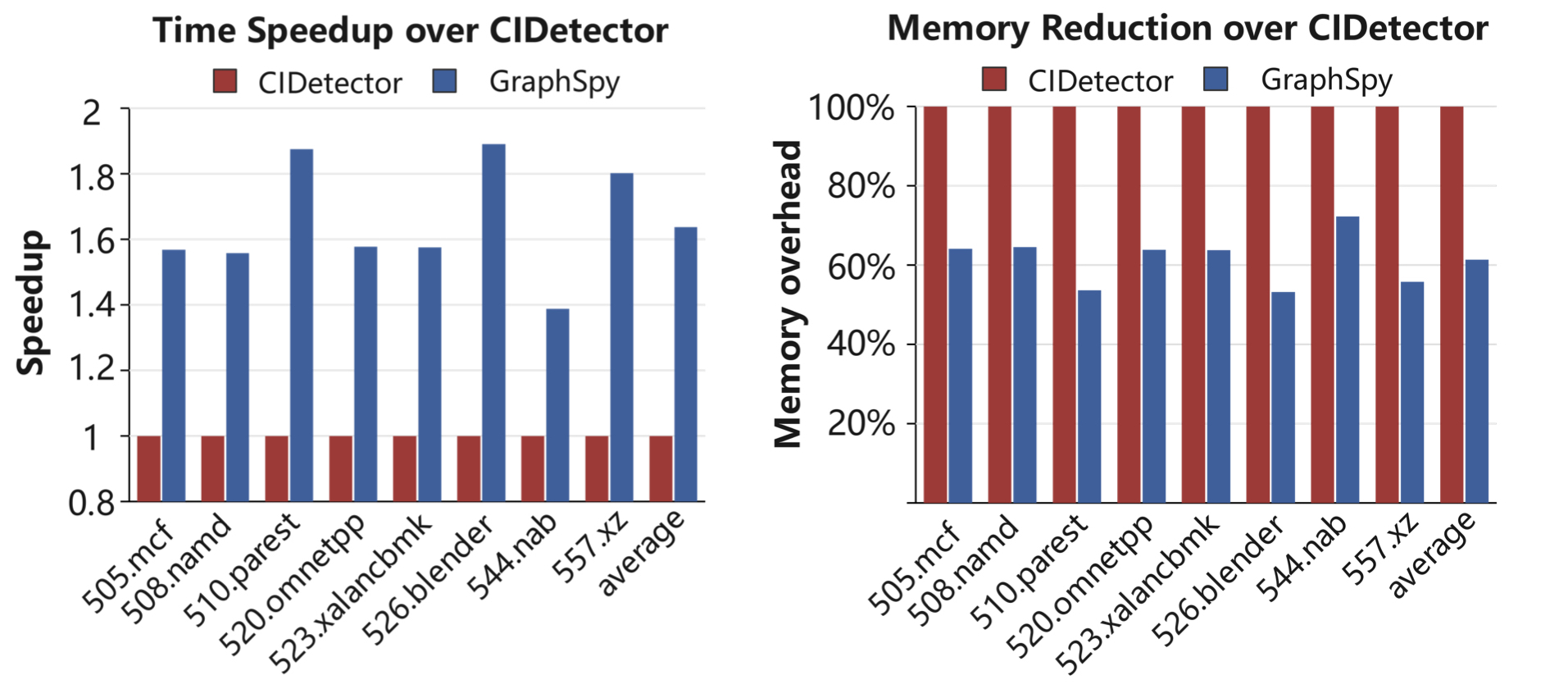}
}
\caption{Time saving and memory reduction over \texttt{CIDetector}.}
\label{fig:time-memory-compare}
\end{figure*}

Taking \emph{ARM-O3} as an example, we measure time saving and memory overhead reduction by \textsc{GraphSpy}. We apply the unified model trained from the hybrid configuration to find out dead stores for all binaries on \emph{ARM-O3}.  \texttt{CIDetector} runs through a binary by filtering those procedures that do not have dead stores. Compared with whole-program monitoring, \textsc{GraphSpy} improves upon \texttt{CIDetector} by an average speedup of $1.65\times$, with a maximum of $1.89\times$, as shown by Fig.~\ref{fig:time-memory-compare}. Since \texttt{CIDetector} incurs up to 150$\times$ time overhead of the native run, the proposed approach decreases the time overhead of \texttt{CIDetector} by a maximum of 71$\times$ of the time of native run (i.e., 150 - 150/1.89). In the mean time, memory cost incurred is reduced as well. Fig.~\ref{fig:time-memory-compare} shows that the proposed approach reduces 40\% of the memory overhead incurred by \texttt{CIDetector}.

\subsection{Model Comparison Results}

\def \gspy {\textsc{GraphSpy}}

Table~\ref{tbl:mdl} compares \gspy\ with alternative models. Clearly, the word2vec model achieves good accuracy of up to 83.05\% when used alone. By contrast, BERT performs worse with accuracy of 57.4\%. It is believed that BERT would potentially perform better. However, its incredibly long training time prevents from being more tuned in a limited time period. Despite competitive accuracy results obtained by only using one model, combining them together absolutely enhances the results further. For example, word2vec + GGNN$_{cfg}$ + Resnet11 enhances the accuracy to 87.03\% on \emph{x86-O3}, as opposed to 79.08\% by only using word2vec. As a result, \textsc{GraphSpy} achieves the highest accuracy of 92.56\% with a combination of word2vec, Resnet, GGNN for CFGs and GGNN for CCTs.

\begin{table}[t]
\centering
\scalebox{0.9}{
\begin{tabular}{|l|c|c|}
\hline
\textbf{Model} &  \textbf{ARM-O3} & \textbf{x86-O3} 
\\\hline\hline
word2vec               & 0.7640/0.8305  & 0.7105/0.7908 \\
BERT                   & 0.6133/0.5740  & 0.6910/0.4580 \\\hline
CNN3                   & 0.4026/0.8402  & 0.3820/0.8446 \\ 
Resnet7                & 0.4712/0.8067  & 0.3698/0.8469 \\ 
Resnet11               & 0.5706/0.7910  & 0.4549/0.8071 \\\hline 
w2v + GGNN$_{cfg}$             & 0.8756/0.7763  & 0.7664/0.7969 \\
BERT + GGNN$_{cfg}$            & 0.5489/0.7610  & 0.3066/0.7940 \\
w2v + GGNN$_{cfg}$ + Res11     & 0.8198/0.8246  & 0.7584/0.8234 \\
BERT + GGNN$_{cfg}$ + Res11    & 0.4272/0.8428  & 0.3333/0.8703 \\\hline 
\textsc{GraphSpy}      & \textbf{0.9070/0.9256}  & \textbf{0.8415/0.8668 }\\\hline 
\end{tabular}
}
\caption{Comparison among different models.}
\label{tbl:mdl}
\end{table}

\section{Related Work}
\textbf{GNNs for program embedding.} GNNs have attracted increasing attention to program representation, partly because many graph structures implicitly existing inside a program code, e.g., abstract syntax tree (AST), CFG and data flow graph, make GNNs highly applicable for program embedding~\cite{Allamanis:corr17}. Prior studies ~\cite{Lu+:corr19,Shi+:ICLR20} use the intermediate representation (IR) or AST to construct CFGs to feed to a gated graph neural network for program classification and data prediction tasks. Due to the inherent difference between syntax and semantics, models learned from static code can be imprecise at capturing semantic properties~\cite{Wang+:ICLR18}. Our model constructs CFGs from the assembly code for a representation of program structure information. By it, architecture specific details can be embedded by using instructions rather than generic intermediate representation.

Wang et al.~\cite{Wang+:ICLR18,WangS:PLDI20} embeds programs from dynamic execution traces or symbolic execution traces, which capture accurate program semantics, thus offering benefits that reason over syntactic representations. However, the quality of the model requires heavy execution profiling. Our dynamic model embeds dynamic program states from calling context trees and sampled memory addresses, which is very lightweight but brings sufficient semantics. Furthermore, our model uses convolutional neural networks to embed relative positional information of basic blocks in CFGs, which is quite unique.


Shi et al.~\cite{Shi+:ICLR20} builds graphs by taking instructions as nodes and value dependency between instructions as edges, and also takes program snapshots of memory values for dynamic embedding. However, it involves no any inter-procedure information and incurs overly much memory and computation overhead.

There are also several works learning from program representation to conduct program repair~\cite{Wang+:ICLR18,WangS:PLDI20}, bug detection~\cite{Dinella+:ICLR20}, program classification~\cite{Lu+:corr19}, cache replacement~\cite{LiG:arxiv20,Liu+:ICML20}, heap memory wise program verification~\cite{Li+:ICLR16}, and code similarity detection~\cite{Yu+:AAAI20}. Our work is the first to apply GNNs for dead store detection.





\vspace{+.2em}
\noindent
\textbf{Memory wastage analysis.} Usage analysis and wastage analysis are two sister problems. Traditional performance analysis tools, such as \emph{gprof}~\cite{Graham+:CC82}, \emph{HPCToolkit}~\cite{Adhianto+:10}, and \emph{vTune}~\cite{VTune:URL} focus on the former, i.e., attributing running time of a program to code structures at various granularities. In parallel, wastage analysis tells how many computation and memory operations are unnecessary, for example, dead store detection~\cite{ChabbiM:CGO12}, redundant load checking~\cite{Su+:ICSE19}, run-time value numbering~\cite{Wen+:ASPLOS18} and value locality exploration~\cite{Wen+:ASPLOS17}. Unfortunately, existing checking tools have limited applicability because of as high as up to 150$\times$ overhead. Our work focuses upon dead store detection and improves the high overhead by a great margin.

\section{Conclusion}
In this paper, we have presented a new learning aided approach namely \gspy\ built upon graph neural networks to detect dead stores with reduced overhead. As far as we know, it is the first work that applies GNNs for dead store detection. We have designed a novel embedding approach that embeds intra- and inter-procedural structure semantics and dynamic memory value semantics to enhance the precision of detection. To detect dead stores across platforms and compilation options, we have presented a CNN-based approach for embedding relative positional information. Upon the evaluation over a set of real-world programs, results turn out that \gspy\ achieves as high as 90\% of accuracy.



\bibliography{all}

\end{document}